\documentclass{article}

\usepackage{amsmath}
\usepackage{amssymb}
\usepackage{graphicx}
\usepackage{placeins}
\usepackage{wrapfig}
\usepackage{float}
\usepackage[table]{xcolor}

\usepackage[preprint]{corl_2026} 

\newlength{\origtextfloatsep}
\newlength{\origfloatsep}
\newlength{\origintextsep}
\newlength{\origabovedisplayskip}
\newlength{\origbelowdisplayskip}
\newlength{\origabovedisplayshortskip}
\newlength{\origbelowdisplayshortskip}
\newcommand{\compactbodyspacing}{%
  \setlength{\abovedisplayskip}{0.55em plus 0.15em minus 0.15em}%
  \setlength{\belowdisplayskip}{0.55em plus 0.15em minus 0.15em}%
  \setlength{\abovedisplayshortskip}{0.35em plus 0.1em minus 0.1em}%
  \setlength{\belowdisplayshortskip}{0.35em plus 0.1em minus 0.1em}%
}

\makeatletter
\let\origsection\section
\let\origsubsection\subsection
\newcommand{\compactbodyheadings}{%
  \renewcommand{\section}{%
    \@startsection{section}{1}{\z@}%
                  {-1.5ex \@plus -0.4ex \@minus -0.2ex}%
                  {1.0ex \@plus 0.2ex \@minus 0.15ex}%
                  {\large\bf\raggedright}%
  }%
  \renewcommand{\subsection}{%
    \@startsection{subsection}{2}{\z@}%
                  {-0.9ex \@plus -0.3ex \@minus -0.15ex}%
                  {0.3ex \@plus 0.1ex}%
                  {\normalsize\bf\raggedright}%
  }%
}
\newcommand{\resetbodyheadings}{%
  \let\section\origsection
  \let\subsection\origsubsection
}
\renewcommand{\paragraph}{%
  \@startsection{paragraph}{4}{\z@}%
                {0.35ex \@plus 0.15ex \@minus 0.1ex}%
                {-1em}%
                {\normalsize\bf}%
}
\makeatother

\definecolor{methodblue}{RGB}{232,234,255}
\newcommand{\methodrow}{\rowcolor{methodblue}}
\newcommand{\taskname}[1]{{\footnotesize\textit{#1}}}

\title{Sparse2Act: Learning Action-Aligned Sparse 3D Representations 
for Cross-Domain Robot Manipulation}

\author{
  Yu Guo$^{1,*}$, Chang Yu$^{1,*}$, Siyu Ma$^{1,2,*}$, \\
  \textbf{Yunuo Chen$^{1}$, Yin Yang$^{3}$,
  Ying Nian Wu$^{1}$, Chenfanfu Jiang$^{1}$} \\
  $^{1}$University of California, Los Angeles, \\
  $^{2}$University of California, San Diego, \\
  $^{3}$University of Utah \\
  {\normalfont\footnotesize * Equal contributions.}
}

\begin{document}

\compactbodyspacing
\compactbodyheadings

\maketitle
\vspace{-2.5em}
\begin{abstract}
Explicit 3D representations are attractive for manipulation because they expose
object shape, workspace geometry, and robot--object relations in metric
coordinates. However, sparse 3D encoders are often learned through downstream task
objectives, tying the representation to a particular data distribution, policy
architecture, and action parameterization. We introduce \textbf{Sparse2Act}, an
observation--action alignment framework for pretraining sparse point-cloud
encoders. The key idea is to use task-space end-effector actions as geometric
supervision: masked sparse 3D tokens are trained to organize scene features
around the workspace motion paired with the observation. After pretraining, only
the encoder initialization is reused by downstream policies, allowing them to retain their
own architectures and action spaces, including joint-space commands. On the
LIBERO-10 benchmark, our method achieves $86.9\%$ average success after 500 fine-tuning
steps. The same pretrained encoder supports LIBERO-to-Meta-World cross-domain
transfer, achieving $73.4\%$ average success on the Meta-World-5 benchmark. Ablations on the objective and decoder capacity show that the gains come from the masked
action-alignment signal and remain useful across downstream action decoders.
In real-world experiments, simulation pretraining followed by limited real-data fine-tuning achieves an average success rate of $72.5\%$ across four tasks, demonstrating effective sim-to-real transfer. These results suggest that robot actions can provide
compact geometric supervision for reusable sparse 3D representations.
Project page: \url{https://sparse2act.github.io/Sparse2Act/}.
\end{abstract}


\section{Introduction}

\begin{figure}[t]
    \centering
    \includegraphics[width=1\linewidth]{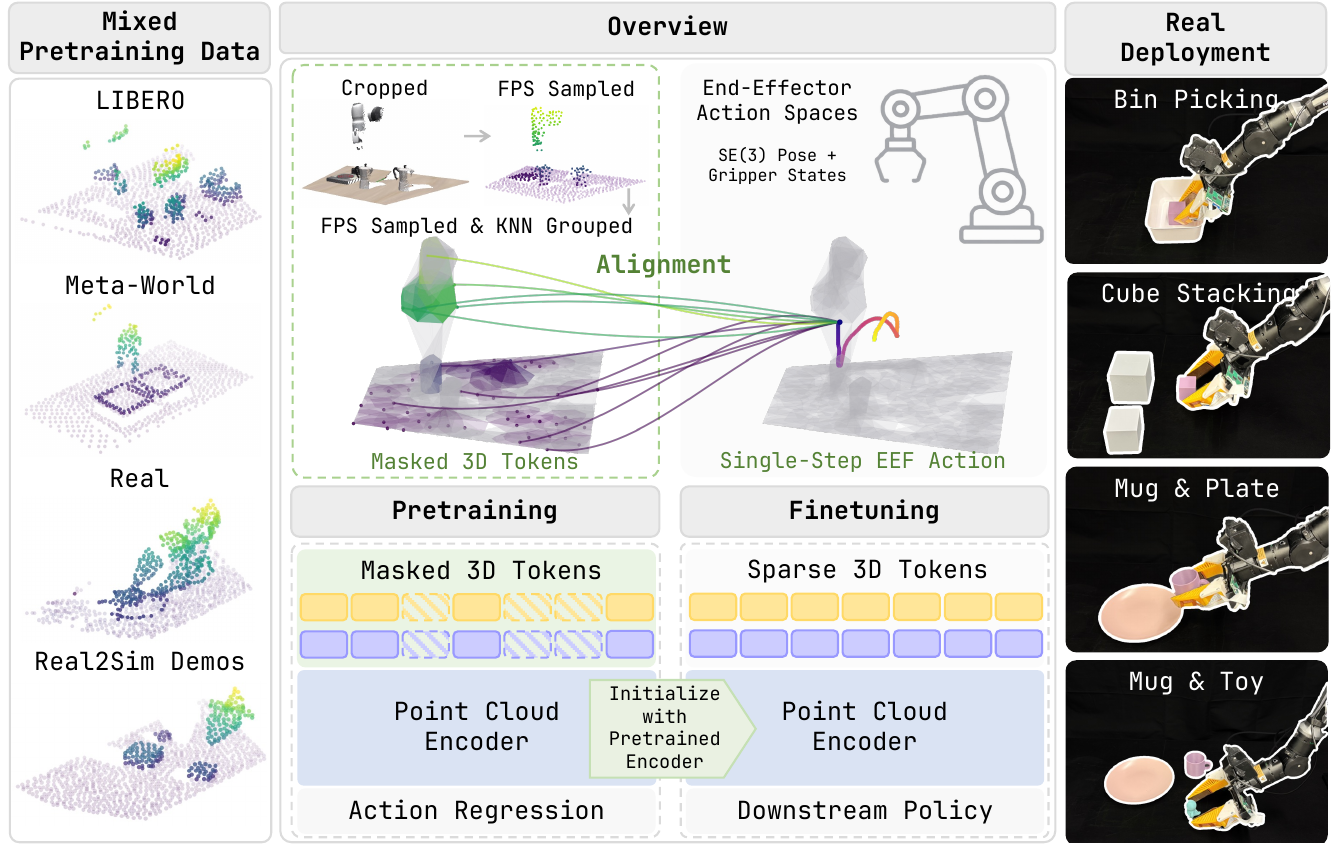}
    \vspace{-0.7em}
    \caption{
    \textbf{Observation--action space alignment for sparse 3D policy pretraining.}
    Sparse 3D observations and end-effector motions share a metric workspace, enabling task-space actions to supervise geometric feature learning. Left: mixed sim and real-to-sim data provide point-cloud–action pairs. Middle: masked sparse 3D tokens are encoded with action supervision, yielding an encoder reused by downstream policies. Right: the pretrained encoder enables cross-domain and sim-to-real transfer for manipulation tasks.
    }
    \label{fig:teaser}
\end{figure}

Explicit 3D representations give manipulation policies direct access to metric
object shape, spatial relations, and robot-object geometry. Recent sparse 3D
policies realize this advantage using point-based~\cite{ze3DDiffusionPolicy2024,chenPolarNet3DPoint2023},
voxel-based~\cite{shridharPerceiverActorMultiTaskTransformer2023,wangRISE3DPerception2024},
or sparse-token-based~\cite{gervetAct3D3DFeature2023,ke3DDiffuserActor2025}
encoders whose features are consumed by downstream action models. In most
policy-learning pipelines, these encoders are co-trained through the downstream
task objective, so control-relevant 3D features are learned under the same data distribution,
controller, and action parameterization used for the final policy. A natural
complement is to shape the encoder before this downstream coupling,
as prior work on robot representation learning has shown that pretrained features can provide a strong initialization that improves
data efficiency and transfer~\cite{nairR3MUniversalVisual2023,maVIPUniversalVisual2023,radosavovicRealWorldRobotLearning2023,radosavovicRobotLearningSensorimotor2023,qian3DMVP3DMultiview2025}.

The pretraining objective therefore matters because it determines which features
the encoder learns. Existing
objectives induce different useful structures: masked reconstruction emphasizes
recoverable 3D content~\cite{yuPointBERTPreTraining3D2022,pangMaskedAutoencodersPoint2022},
dynamics and future prediction emphasize temporal evolution~\cite{cuiDynaMoInDomainDynamics2024a,hou4DVisualPretraining,liangBootstrapDynamicAware3D2025,yuanFastWAMWorldAction2026,huangPointWorldScaling3D2026},
and contrastive, value-based, semantic, or multimodal signals emphasize invariance, alignment, and consistency~\cite{nairR3MUniversalVisual2023,maVIPUniversalVisual2023,chenSUGARPretraining3D2024,zhuSPA3DSpatialAwareness2024,qian3DMVP3DMultiview2025,liuCLAMPContrastiveLearning2026a}.
These targets can all support control in different ways, but they leave open a broader
representation question: whether 3D end-effector-space actions can be used during pretraining 
to shape a sparse 3D encoder that operates in the same 3D workspace.

The key observation follows from this \emph{shared 3D workspace}: task-space actions provide
the encoder with a geometric alignment signal between the current 3D state and
robot motion~\cite{gervetAct3D3DFeature2023,ze3DDiffusionPolicy2024,huangPointWorldScaling3D2026}.
With masked sparse tokens, this supervision is applied through the encoded scene
representation, encouraging features that organize visible geometry in relation
to controllable motion. We instantiate this idea as \textbf{Sparse2Act}, an encoder-level pretraining
framework for observation--action alignment. During
pretraining, our encoder learns from masked point-cloud observations and
task-space action supervision; during downstream learning, policies reuse the pretrained encoder weights as initialization while retaining their own action parameterizations.

Our contributions are threefold: (1) an observation--action alignment objective
for native sparse 3D encoder pretraining, where masked point-cloud features are
supervised by task-space actions to acquire an action-aligned geometric prior;
(2) an encoder-transfer pipeline that uses the pretrained sparse 3D encoder to bootstrap downstream policy learning across changes in task, domain, and action space; and (3) evaluations spanning in-domain adaptation, LIBERO-to-Meta-World transfer, data-efficient fine-tuning, and real-robot deployment. 
Our pretraining framework achieves an average success rate of $86.9\%$ on LIBERO-10 and $73.4\%$ in LIBERO-10-to-Meta-World-5 transfer. Furthermore, simulation pretraining enhances sim-to-real performance when combined with real-data fine-tuning, yielding an average success rate of $72.5\%$ across four real-world tasks.

\section{Related Work}

\paragraph{3D Manipulation Policies.}
Image-based policies typically learn task-relevant geometry implicitly from camera-view
observations~\cite{brohanRT1RoboticsTransformer2023,chiDiffusionPolicyVisuomotor2023,kimOpenVLAOpenSourceVisionLanguageAction2024}.
3D representations expose workspace coordinates, object shape, and contact
geometry more explicitly. This has made 3D perception useful for grasping and
geometry-driven action proposals~\cite{tenPasGraspPoseDetection2017,liangPointNetGPDDetecting2019,fangGraspNet1BillionLargeScale2020,sundermeyerContactGraspNetEfficient2021},
articulated-object interaction and object-centric manipulation~\cite{moWhere2ActPixelsActions2021,eisnerFlowBot3DLearning2022,zhuLearningGeneralizableManipulation2023},
as well as reinforcement learning (RL) and sim-to-real manipulation~\cite{huangGeneralizationDexterousManipulation2021,baoDexArtBenchmarkingGeneralizable2023,xuUniDexGraspUniversalRobotic2023,qin2023dexpoint,yu2025right}.
Other point-cloud manipulation methods use motion-centric targets such as tool
flow or point trajectories to derive actions~\cite{seitaToolFlowNetRoboticManipulation2023,wenAnypointTrajectoryModeling2024}.
Imitation work uses the same geometric prior for action prediction with voxel,
multi-view, or action-map policies~\cite{shridharPerceiverActorMultiTaskTransformer2023,goyalRVTRoboticView2023,gervetAct3D3DFeature2023},
and sparse point-cloud policies~\cite{ze3DDiffusionPolicy2024,wangRISE3DPerception2024,ke3DDiffuserActor2025,chenPolarNet3DPoint2023,haldarPointBridge3D2026}.
Foundation-policy work likewise injects 3D structure into larger or
vision-language-action (VLA) policies~\cite{yangFP33DFoundation2025,quSpatialVLAExploringSpatial2025,liPointVLAInjecting3D2026}.
These works mainly evaluate 3D encoders inside the final policy; we instead
study how to efficiently pretrain the sparse 3D encoder before downstream control.

\paragraph{Pretraining 3D Representations for Control.}
Representation pretraining for robot policies spans several related lines.
Generic 3D methods such as Point-BERT~\cite{yuPointBERTPreTraining3D2022} and
Point-MAE~\cite{pangMaskedAutoencodersPoint2022} learn geometric features
through recognition or reconstruction objectives.
Robotics-oriented 3D pretraining adds scene- and action-relevant supervision,
with SUGAR~\cite{chenSUGARPretraining3D2024} and
SPA~\cite{zhuSPA3DSpatialAwareness2024} using semantic or spatial objectives,
3D-MVP~\cite{qian3DMVP3DMultiview2025} using multi-view geometry,
CLAMP~\cite{liuCLAMPContrastiveLearning2026a} using action-conditioned contrast,
and FVP~\cite{hou4DVisualPretraining} and
AFRO~\cite{liangBootstrapDynamicAware3D2025} using future prediction or
dynamics. Closely related 3D world-modeling work, such as
PointWorld~\cite{huangPointWorldScaling3D2026}, scales action-conditioned scene
prediction through point-flow modeling.
In parallel, control-oriented visual and sensorimotor pretraining uses
time-contrastive, masked, and value-based visual objectives~\cite{nairR3MUniversalVisual2023,maVIPUniversalVisual2023,radosavovicRealWorldRobotLearning2023},
proprioceptive, dynamics, and action objectives~\cite{radosavovicRobotLearningSensorimotor2023,jiangRobotsPretrainRobots2025,cuiDynaMoInDomainDynamics2024a},
and video or world-action objectives~\cite{liUnifiedVideoAction2025a,zhuUnifiedWorldModels2025b,yuanFastWAMWorldAction2026}.
These works suggest that control-relevant pretraining can benefit manipulation policies; however, they typically rely on indirect objectives such as reconstruction, semantic or geometric prediction, dynamics modeling, value learning, or multimodal sensorimotor prediction. In contrast, we investigate whether robot actions can serve as direct, control-grounded supervision for pretraining native sparse 3D encoders.

\section{Method}
\label{sec:method}

\begin{figure}[t]
  \centering
\includegraphics[width=\linewidth]{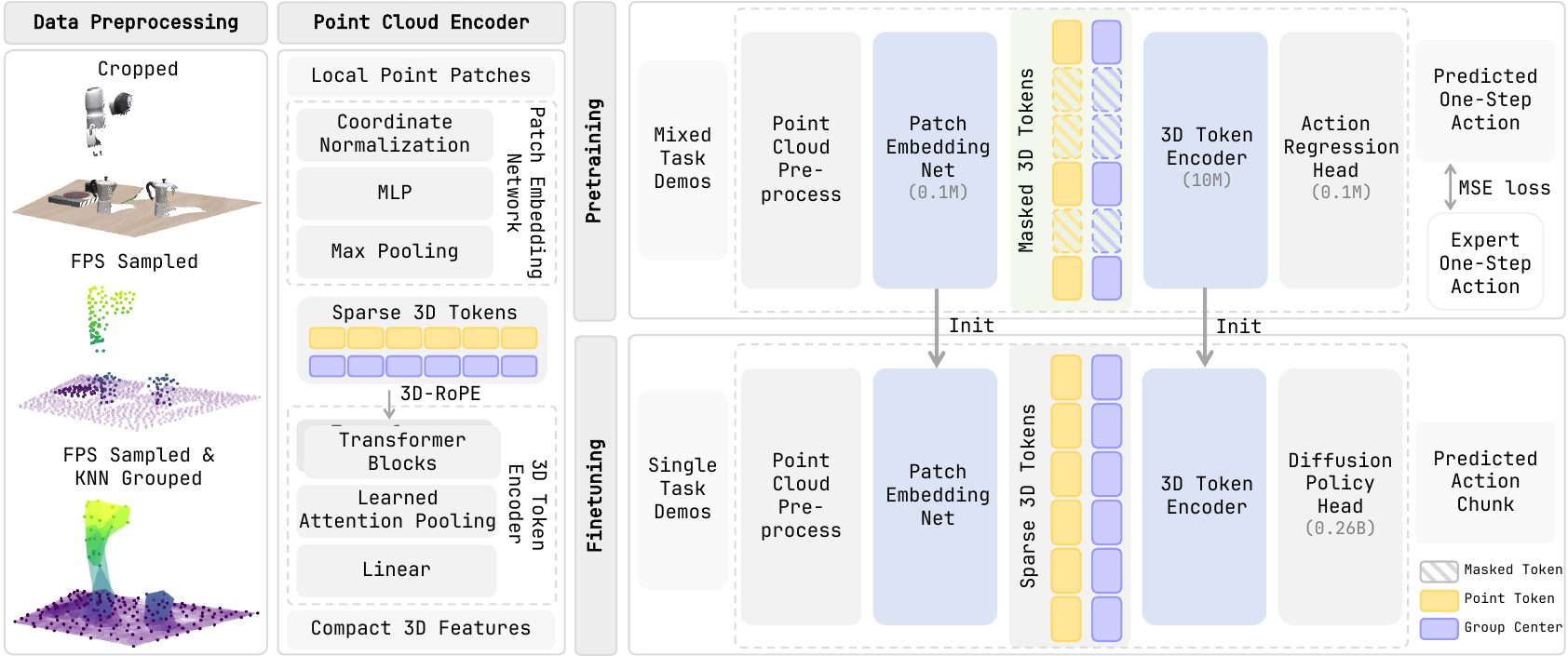}
  \vspace{-0.5em}
  \caption{\textbf{Framework overview.}
  \textit{Left:} raw point clouds are cropped, sampled, and grouped into local
  point patches. \textit{Middle:} local patches are embedded as sparse 3D tokens
  and processed by a 3D token encoder with 3D RoPE. \textit{Right-top:} masked
  sparse 3D tokens are trained with task-space alignment actions, shaping the
  encoder around geometry--motion structure in the shared workspace.
  \textit{Right-bottom:} the pretrained encoder initializes the point-cloud
  branch of a downstream policy, which learns its own deployed action
  parameterization.}
  \label{fig:method-overview}
\end{figure}

Our framework uses 3D-aligned action supervision to pretrain the sparse 3D encoder before
downstream policy learning (Figure~\ref{fig:method-overview}). The key design is
to apply this supervision at the encoder level: masked point-cloud tokens are
supervised by task-space actions, and the resulting encoder initialization is
then reused inside downstream policies. This separation allows pretraining to exploit
the geometric alignment between point clouds and task-space motion, while the
deployed policy remains free to use the controller action space preferred for
control stability~\cite{fengDemystifyingActionSpace2026}. Details of the 3D encoder, masking scheme, positional encoding,
and pretraining setup are provided in Appendix~\ref{app:method-details};
fine-tuning hyperparameters are provided in Appendix~\ref{app:downstream-hparams}.

\subsection{Two-Stage Training Decomposition}

Let $\mathcal{P}$ be a point cloud, $\mathcal{T}(\mathcal{P})$ its sparse 3D
token set constructed below, and $\tilde{\mathcal{T}}(\mathcal{P})$ the masked
token set. We first train an encoder $E_\phi$ and auxiliary alignment head
$h_\psi$ on point-cloud--action pairs
$\mathcal{D}_{\mathrm{pre}}=\{(\mathcal{P},\mathbf{a}^{\mathrm{align}})\}$, where
$\mathbf{a}^{\mathrm{align}}$ denotes the pretraining alignment action:
\begin{equation}
  (\phi^\star,\psi^\star)
  =
  \arg\min_{\phi,\psi}
  \mathbb{E}_{(\mathcal{P},\mathbf{a}^{\mathrm{align}})
  \sim\mathcal{D}_{\mathrm{pre}}}
  \left[
    \|h_\psi(E_\phi(\tilde{\mathcal{T}}(\mathcal{P})))
    - \mathbf{a}^{\mathrm{align}}\|_2^2
  \right].
\end{equation}
After pretraining, $E_{\phi^\star}$ is used to initialize the point-cloud encoder in a
downstream policy. The downstream policy head $\pi_\theta$, with parameters
$\theta$, maps the encoded point-cloud latent and proprioception
$\mathbf{s}$ to the deployed action $\mathbf{a}$. The policy is then trained on
$\mathcal{D}_{\mathrm{down}}=\{(\mathcal{P},\mathbf{s},\mathbf{a})\}$ with a
behavior-cloning loss $\mathcal{L}_{\mathrm{BC}}$:
\begin{equation}
  \min_{\phi,\theta}
  \mathbb{E}_{(\mathcal{P},\mathbf{s},\mathbf{a})
  \sim\mathcal{D}_{\mathrm{down}}}
  \left[
    \mathcal{L}_{\mathrm{BC}}\bigl(
      \pi_\theta(E_\phi(\mathcal{T}(\mathcal{P})),\mathbf{s}),\mathbf{a}
    \bigr)
  \right],
  \qquad \phi \leftarrow \phi^\star.
\end{equation}
Thus, $\mathbf{a}^{\mathrm{align}}$ supervises encoder pretraining, while
$\mathbf{a}$ serves as the downstream control target. The two stages share only
the sparse 3D encoder, allowing the pretraining data and action space to differ
from those used for downstream control.

\subsection{Sparse 3D Encoder}

\paragraph{Sparse 3D tokens.}
Each observation is a sparse point cloud of $N$ points,
$\mathcal{P}=\{\mathbf{p}_i\}_{i=1}^N$, where
$\mathbf{p}_i \in \mathbb{R}^3$ denotes an $xyz$ coordinate.  We restrict inputs to $xyz$-only point
clouds to center pretraining on the \emph{shared 3D workspace} in which both object geometry and
end-effector motion are expressed. This compact observation interface encourages the encoder to focus on geometry--action alignment rather than appearance-specific cues.

Following the DP3 point-cloud preprocessing pipeline~\cite{ze3DDiffusionPolicy2024},
we first crop the raw 3D point cloud by a scene bounding box and downsample it to a fixed
number of points. To convert the unordered cloud into structured local geometric units, we
adopt the Point-BERT patch construction design~\cite{yuPointBERTPreTraining3D2022}:
patch centers are selected by farthest point sampling and local patches are
formed using $k$-nearest-neighbor search. A shared point-patch embedding network
then maps each patch to a continuous embedding.
The resulting patches $\{\mathcal{P}_j\}_{j=1}^M$ define sparse 3D tokens
$\mathcal{T}(\mathcal{P})=\{(\mathbf{t}_j,\mathbf{c}_j)\}_{j=1}^M$, where
$\mathbf{t}_j$ is the patch embedding and $\mathbf{c}_j \in \mathbb{R}^3$ its
center. A Transformer encoder $E_\phi$ maps these tokens to an observation
latent $\mathbf{z} = E_\phi(\mathcal{T}(\mathcal{P}))$.

\paragraph{3D rotary encoding.}
While patch embeddings encode local geometry, 3D Rotary Positional Embeddings (3D RoPE) inject patch workspace coordinates into the attention mechanism~\cite{suRoFormerEnhancedTransformer2024}.
Let $\bar{\mathbf{c}}_j=(\bar{x}_j,\bar{y}_j,\bar{z}_j)$ denote the normalized center
of patch $\mathcal{P}_j$. For each attention head, the query and key channels are partitioned into three coordinate-aligned groups,
$\mathbf{q}_j=[\mathbf{q}_j^x;\mathbf{q}_j^y;\mathbf{q}_j^z]$ and
$\mathbf{k}_j=[\mathbf{k}_j^x;\mathbf{k}_j^y;\mathbf{k}_j^z]$. For
$\mathbf{v}=[\mathbf{v}^x;\mathbf{v}^y;\mathbf{v}^z]$, the axis-wise RoPE
transform is defined as
\begin{equation}
  \rho(\mathbf{v},\bar{\mathbf{c}}_j) =
  \left[
  \mathbf{R}(\bar{x}_j)\mathbf{v}^x;\,
  \mathbf{R}(\bar{y}_j)\mathbf{v}^y;\,
  \mathbf{R}(\bar{z}_j)\mathbf{v}^z
  \right].
\end{equation}
Here, $\mathbf{R}(u)$ applies the standard RoPE rotation to channel pair $r$,
\begin{equation}
  \begin{bmatrix}
    v_{2r}' \\
    v_{2r+1}'
  \end{bmatrix}
  =
  \begin{bmatrix}
    \cos(u\omega_r) & -\sin(u\omega_r) \\
    \sin(u\omega_r) &  \cos(u\omega_r)
  \end{bmatrix}
  \begin{bmatrix}
    v_{2r} \\
    v_{2r+1}
  \end{bmatrix},
\end{equation}
where $\{\omega_r\}$ denotes the standard frequency schedule. Attention uses the rotated
queries and keys:
\begin{equation}
  \alpha_{ij}
  =
  \mathrm{softmax}_{j}
  \left(
    \frac{\rho(\mathbf{q}_i,\bar{\mathbf{c}}_i)^\top
    \rho(\mathbf{k}_j,\bar{\mathbf{c}}_j)}
    {\sqrt{d_h}}
  \right),
\end{equation}
where $d_h$ is the per-head feature dimension.

\subsection{3D Action-Aligned Masked Pretraining}

Pretraining aligns masked sparse 3D observations with task-space motion in the
shared workspace. We randomly mask a subset of token embeddings before the
Transformer encoder. Unlike masked autoencoding objectives that reconstruct visual or
point-cloud content~\cite{heMaskedAutoencodersAre2022,pangMaskedAutoencodersPoint2022},
the masked tokens are not treated as reconstruction targets; instead, the alignment loss is applied
to the encoded visible geometry. Let $m_j \in \{0,1\}$ denote the binary mask for
token $\mathbf{t}_j$, where $m_j=1$ indicates a visible token and $m_j=0$ a masked
token. We define $\tilde{\mathbf{t}}_j=m_j\mathbf{t}_j$ and write the masked token
set as
$\tilde{\mathcal{T}}(\mathcal{P})=\{(\tilde{\mathbf{t}}_j,\mathbf{c}_j)\}_{j=1}^M$.
The alignment head predicts the task-space alignment action from the masked
scene representation:
\begin{equation}
  \hat{\mathbf{a}}^{\mathrm{align}}
  = h_\psi(E_\phi(\tilde{\mathcal{T}}(\mathcal{P}))).
\end{equation}
The model is trained with the objective
$\mathcal{L}_{\mathrm{pretrain}}=
\|\hat{\mathbf{a}}^{\mathrm{align}}-\mathbf{a}^{\mathrm{align}}\|_2^2$.
Masking forces the alignment problem to depend on scene-level spatial relations
among visible tokens, encouraging the encoder to organize sparse geometric structure around
3D end-effector motion within the shared workspace.

\subsection{Downstream Policy Initialization}

For downstream learning, we initialize the point-cloud encoder with
$E_{\phi^\star}$ and attach a task-specific policy head that fuses the encoded
latent $\mathbf{z}$ with proprioceptive input $\mathbf{s}$ to produce the control
action $\mathbf{a}$. The 3D alignment head has already fulfilled its role by injecting task-space alignment signals into $E_\phi$; leaving it out keeps the controller interface and action parameterization native for control.

\begin{figure}[t]
  \centering
  \includegraphics[width=\linewidth]{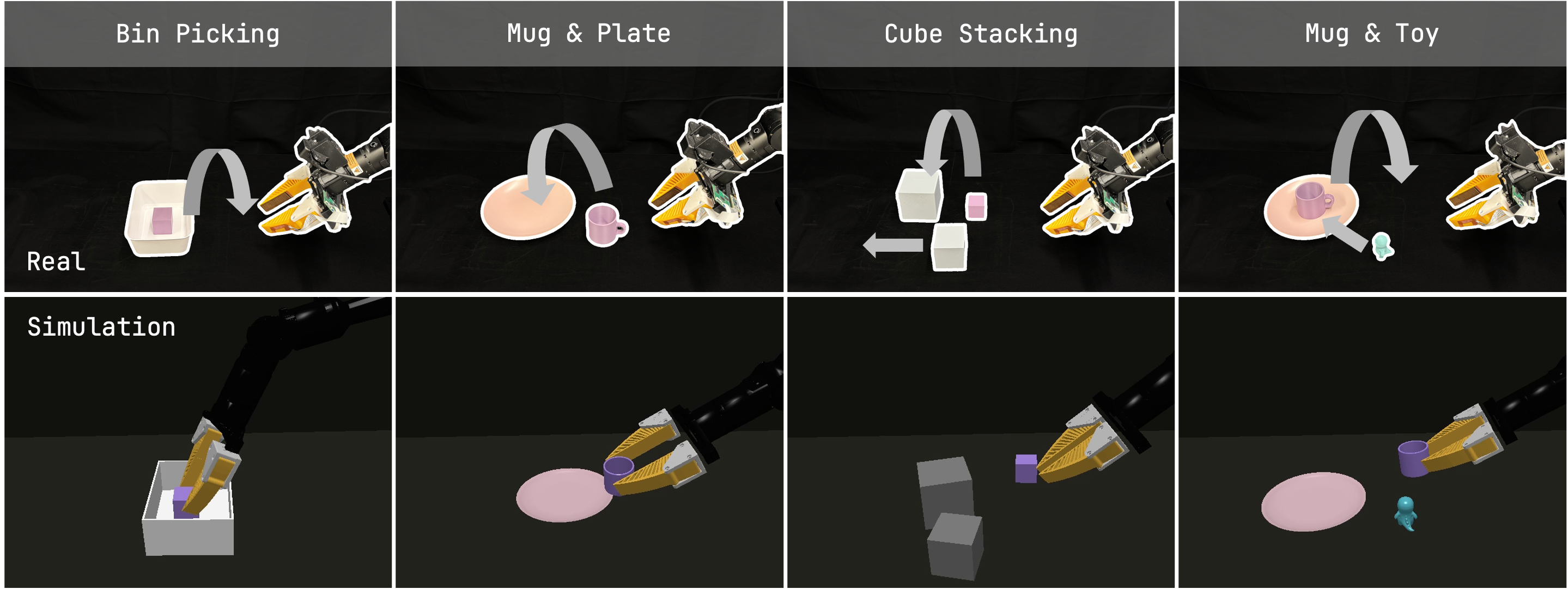}
  \vspace{-1.2em}
  \caption{Real-robot settings and real-to-sim simulation digital twins for
  our tasks.}
  \label{fig:real-robot-qualitative}
\end{figure}

\section{Experiments}

Our experiments evaluate whether the pretrained sparse 3D encoder improves downstream
policy learning, supports cross-domain transfer beyond the pretraining domain,
and remains effective under limited data, simplified action decoders, and sim-to-real
fine-tuning. We first assess in-domain adaptation on LIBERO-10~\cite{liuLIBEROBenchmarkingKnowledge2023} and cross-domain transfer from LIBERO to Meta-World-5~\cite{yuMetaWorldBenchmarkEvaluation2020} (Sec.~\ref{sec:main-results}).
We then perform ablations on data efficiency (Sec.~\ref{sec:data-eff}), training objective (Sec.~\ref{sec:obj-ablation}), and decoder capacity (Sec.~\ref{sec:policy-head}) to attribute performance gains to the pretrained representation.
We then evaluate sim-to-real transfer by fine-tuning
sim-pretrained encoders with limited real-robot data (Sec.~\ref{sec:real-robot}).

\subsection{Experimental Setup}
\label{sec:exp-setup}

We evaluate on LIBERO-10 and Meta-World-5, a five-task subset of Meta-World consisting of \taskname{Bin Picking}, \taskname{Pick Out of Hole}, \taskname{Push}, \taskname{Soccer}, and \taskname{Stick Pull}. For consistency, all sparse 3D policy experiments use the same DP3 point-cloud preprocessing pipeline~\cite{ze3DDiffusionPolicy2024}: depth observations are unprojected to point clouds, cropped to the task workspace, and downsampled to 1024 points via farthest point sampling. Checkpoint selection, evaluation protocols, and sources of reported numbers are detailed in Appendix~\ref{app:protocol}.
We compare against DP3~\cite{ze3DDiffusionPolicy2024} as the primary sparse point-cloud policy baseline; 3D pretraining methods FVP~\cite{hou4DVisualPretraining} and AFRO~\cite{liangBootstrapDynamicAware3D2025}; and vision-language-action (VLA) approaches SpatialVLA~\cite{quSpatialVLAExploringSpatial2025} and $\pi_0$~\cite{blackP0VisionLanguageActionFlow2025}.

\subsection{Main Benchmark Results}
\label{sec:main-results}

\begin{figure}[!t]
  \centering
  \includegraphics[width=0.64\linewidth]{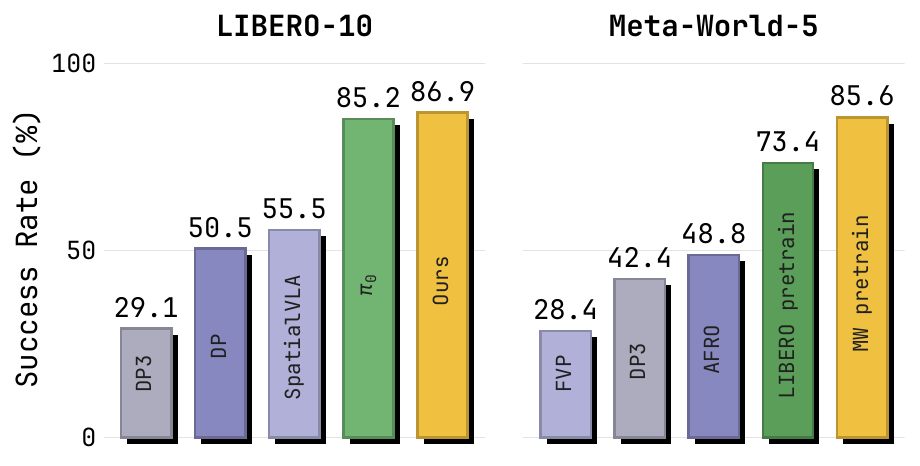}
  \vspace{-0.8em}
  \caption{Success rates on LIBERO-10~\cite{liuLIBEROBenchmarkingKnowledge2023}
and Meta-World-5~\cite{yuMetaWorldBenchmarkEvaluation2020}. Our approach improves
in-domain adaptation and enables LIBERO-10-to-Meta-World-5 cross-domain transfer.}
  \label{fig:main-results}
\end{figure}

Figure~\ref{fig:main-results} evaluates our method under both in-domain and cross-domain settings. On LIBERO-10, the in-domain pretrained encoder achieves an average success rate of 86.9\%, compared to 29.1\% for DP3 trained from scratch under the same point-cloud interface. It also matches the reported performance of $\pi_0$ (85.2\%) and surpasses SpatialVLA (55.5\%), while relying on a sparse 3D policy rather than a large VLA backbone.

On Meta-World-5, the in-domain pretrained encoder achieves 85.6\%, outperforming FVP’s future-prediction pretraining (28.4\%), DP3 trained from scratch (42.4\%), and AFRO’s dynamics-aware 3D pretraining (48.8\%). These comparisons isolate the effect of pretraining from the choice of objective: while predictive and dynamics-based 3D objectives provide useful structure, they do not explicitly align current sparse geometry with task-space motion for downstream control.
When pretrained on LIBERO-10 and fine-tuned on Meta-World-5, our encoder still achieves 73.4\% success, retaining much of its in-domain performance despite the shift in task suite and scene distribution. Full per-task statistics are provided in Table~\ref{tab:e1-libero10-per-task} and Table~\ref{tab:metaworld5-full}.

\subsection{Data Efficiency and Scaling}
\label{sec:data-eff}

\begin{figure}[!htbp]
  \centering
  \includegraphics[width=\linewidth]{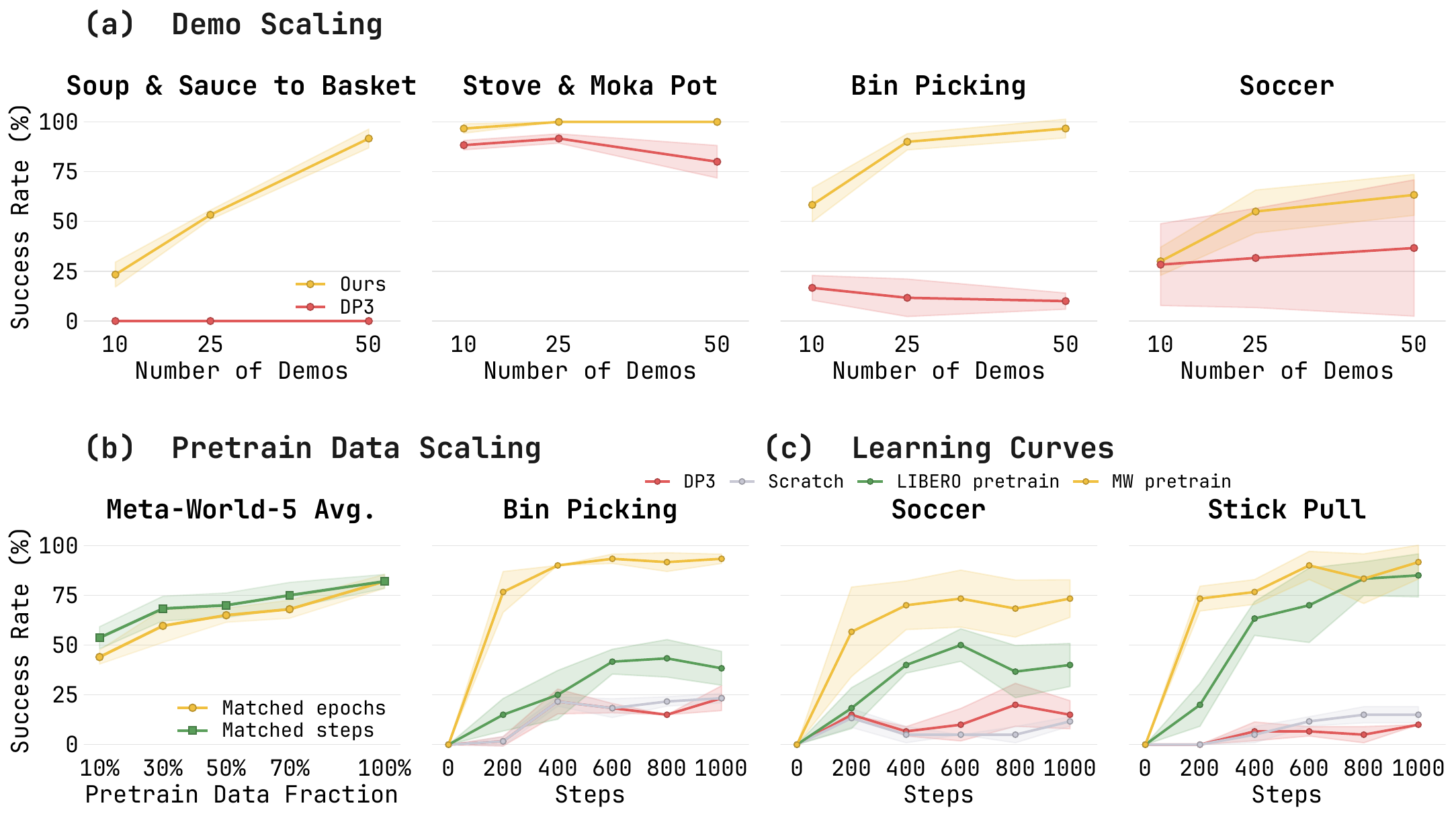}
  \vspace{-1.5em}
  \caption{%
    \textbf{Data efficiency and scaling behavior.} (a) Pretraining improves success across
    downstream demonstration budgets, with the largest gains on harder tasks.
    (b) Performance increases with pretraining coverage under both matched- step
    and epoch schedules. (c) In-domain pretraining converges fastest,
    while LIBERO-to-Meta-World transfer remains consistently above scratch.}
  \label{fig:data-efficiency}
\end{figure}

\paragraph{Demonstration-budget scaling.}
Figure~\ref{fig:data-efficiency}(a) evaluates downstream demonstration budgets
on representative LIBERO-10 and Meta-World-5 tasks. Pretraining is already usable
with 10 demonstrations and continues to improve as more task-specific data is
added on the harder tasks. On easier tasks, performance saturates early,
suggesting that the pretrained encoder mainly reduces the amount of downstream
data needed to reach high success.

\paragraph{Pretraining data scaling.}
Figure~\ref{fig:data-efficiency}(b) varies the pretraining data ratio under two
schedules: one scales pretraining steps with the amount of data, and the other
keeps the step count fixed (Appendix~\ref{app:pretrain-data-scaling}).
Performance improves in both schedules, indicating that additional pretraining
coverage helps beyond simply running more pretraining steps.

\paragraph{Learning curves.}
Figure~\ref{fig:data-efficiency}(c) shows that in-domain pretraining
reaches strong performance within a few hundred fine-tuning steps and is saturated by 600 steps. Cross-domain initialization from LIBERO
improves over scratch but requires longer fine-tuning to approach the
in-domain curves. The DP3-style scratch baseline remains low after 1,000 training steps, suggesting that the downstream data budget alone is insufficient to shape a comparable sparse 3D representation from scratch.

\subsection{Pretraining Objective Ablations}
\label{sec:obj-ablation}

We ablate the pretraining signal to separate the value of initialization from
the value of 3D action alignment. On Meta-World-5, all variants use 50 downstream
demonstrations per task and the same 2{,}000-step fine-tuning schedule. We
compare training from scratch, unmasked action alignment, masked reconstruction,
and masked action alignment.

\begin{table}[t]
  \centering
  \begin{minipage}[t]{0.43\linewidth}
    \centering
    \small
    \caption{Pretraining objective ablation on Meta-World-5.}
    \label{tab:objective-ablation}
    \begin{tabular}{lc}
      \hline
      Pretraining objective & Success \\
      \hline
      None & 19.0 $\pm$ 4.6 \\
      Action only & 50.7 $\pm$ 5.5 \\
      Reconstruction only & 55.3 $\pm$ 8.0 \\
      \methodrow
      Masked input + action & \textbf{82.0 $\pm$ 3.5} \\
      \hline
    \end{tabular}
  \end{minipage}
  \hfill
  \begin{minipage}[t]{0.53\linewidth}
    \centering
    \small
    \setlength{\tabcolsep}{3.0pt}
    \caption{Action-decoder capacity ablation on Meta-World-5 at 2{,}000 training steps.}
    \label{tab:policy-simplification}
    \begin{tabular}{lcc}
      \hline
      Action decoder & Scratch & Pretrained \\
      \hline
      DP3 & 18.0 $\pm$ 4.6 & \cellcolor{methodblue}\textbf{82.0 $\pm$ 3.5} \\
      SimpleDP3 & 13.0 $\pm$ 3.6 & \cellcolor{methodblue}75.0 $\pm$ 8.9 \\
      MLP & 12.3 $\pm$ 5.7 & \cellcolor{methodblue}74.7 $\pm$ 3.5 \\
      \hline
    \end{tabular}
  \end{minipage}
  \vspace{0.2em}

  \begin{minipage}[t]{\linewidth}
    \centering
    \small
    \setlength{\tabcolsep}{6.0pt}
    \caption{Single-arm AgileX PiPER real-robot deployment results. Training protocols
    and data sources are detailed in Table~\ref{tab:real-robot-protocols}.}
    \label{tab:real-robot}
    \begin{tabular}{lccccc}
      \hline
      Method & \taskname{Bin Picking} & \taskname{Cube Stacking}
      & \taskname{Mug \& Plate} & \taskname{Mug \& Toy} & Avg. \\
      \hline
      DP3 & 3/10 & 0/10 & 5/10 & 0/10 & 20\% \\
      Ours from scratch & 2/10 & 0/10 & 6/10 & 2/10 & 25\% \\
      \methodrow
      Ours & \textbf{8/10} & \textbf{5/10} & \textbf{9/10} & \textbf{7/10} & \textbf{72.5\%} \\
      \hline
    \end{tabular}
  \end{minipage}
\end{table}

Table~\ref{tab:objective-ablation} shows that both components matter.
Unmasked action alignment and masked reconstruction each improve over scratch,
while masked action alignment offers the largest gain. This suggests that the
benefit is not simply from using actions as labels or from masking as a
regularizer; the combined objective encourages the encoder to connect 3D visible
scene context with task-space 3D motion. Extending the scratch baseline training to
16{,}000 steps raises success from 19.0\% to 28.3\%, still far
below the pretraining variants, suggesting that the gap is not explained by training budget alone.
Full per-task results are in Table~\ref{tab:e4-objective-ablation-full}.

\subsection{Attribution to the Encoder}
\label{sec:policy-head}

If our method makes the 3D latent more directly usable for control, the benefit
should persist when the downstream action decoder is simplified. We test this by
replacing the standard DP3 decoder with SimpleDP3 and an MLP under the same fixed
2{,}000-step training schedule. Table~\ref{tab:policy-simplification} shows that the pretrained
encoder improves all three decoders, and that the simplified decoders retain most
of the gain. Because lower-capacity decoders have less room to compensate for a
weak encoder, their strong performance suggests that most of the improvement is
carried by the pretrained representation rather than the decoder architecture:
SimpleDP3 reaches 75.0\% and the MLP reaches 74.7\%, compared with 82.0\% for
the full DP3 decoder. Parameter statistics and full per-task results are in
Tables~\ref{tab:e6-policy-head-params}
and~\ref{tab:e6-policy-head-simplification-full}.

\subsection{Sim-to-Real Transfer}
\label{sec:real-robot}

We evaluate sim-to-real transfer on the real and simulated settings shown in
Figure~\ref{fig:real-robot-qualitative}, as the strongest domain shift in our study:
the encoder is pretrained in simulation and then adapted to a real single-arm
AgileX PiPER platform with different sensing, dynamics, and controller
conditions. The real-robot suite contains four tasks: \taskname{Bin Picking},
\taskname{Cube Stacking}, \taskname{Mug \& Plate}, and \taskname{Mug \& Toy}.
For each task, we collect 50 demonstrations from real world and 100 demonstrations
from a simulation environment matched to real.

This setting tests whether our action-aligned pretraining pipeline can convert simulated demonstrations into a useful encoder initialization for real-robot policy learning, so that downstream fine-tuning benefits from only a small amount of real data. For matched-data baselines, we co-train the final policy from scratch using the same 100 simulated and 50 real
demonstrations available for each task. Our pretrained variant instead learns
the encoder from simulation first using end-effector actions in the 3D point-cloud
workspace, then fine-tunes with only the 50 real demonstrations using
joint-space actions.
Training and data budgets are matched within each protocol.

Table~\ref{tab:real-robot} shows that, under this comparison,
the way simulated demonstrations are used matters substantially for cross-domain transfer. 
Co-training reaches 20\% with DP3 and 25\% with our architecture from scratch, while simulation pretraining followed by real fine-tuning reaches 72.5\%. 
Co-training requires the policy to fit sim and real distributions jointly, where the small real subset can be dominated by the larger simulated batch and the action decoder can fit simulation-specific behavior that transfers poorly.
Pretraining the encoder separates 3D representation learning from target-controller fitting, leaving the real fine-tuning step free to specialize on the target distribution.

\section{Conclusion}

We introduce \textbf{Sparse2Act}, an action-aligned pretraining framework for learning transferable sparse 3D representations. The central idea is to use task-space actions as geometric supervision for a masked point-cloud encoder prior to downstream policy learning. Since downstream training reuses only the encoder initialization, the policy retains its own architecture and action parameterization, including joint-space commands, while benefiting from a representation shaped in the shared metric workspace.

Across simulated benchmarks and real-robot experiments, this encoder initialization improves in-domain adaptation, cross-domain transfer, data-efficient fine-tuning, and sim-to-real transfer. Ablations on objective design and decoder capacity further show that the gains stem from the pretraining signal and remain robust across downstream action decoders. Overall, these results support a representation-level view of robot actions: in sparse 3D manipulation, actions can serve as supervision for learning reusable 3D encoders prior to downstream control.

\section{Limitations}

\textit{Policy and observation scope.} We study a controlled setting centered on
current sparse point-cloud observations and encoder initialization. This keeps
the role of the 3D representation clear, but leaves larger policy architectures,
multi-frame context, language conditioning, and complementary sensing modalities
such as touch for future study.
\vspace{-0.5em}

\textit{Task and domain coverage.} The experiments focus on manipulation tasks
with metric point-cloud observations and teleoperation-style data. Larger
embodiment changes, deformable objects, cluttered workspaces, and longer-horizon
task structure may require combining action alignment with additional
representation signals.
\vspace{-0.5em}

\textit{Real-world scale.} The real-robot evaluation uses one platform and four
tasks. Scaling to broader real-world pretraining data and more diverse task families remains promising future work.

\bibliography{Corl26}

\newpage
\appendix
\section{Appendix}

\subsection{Method and Implementation Details}
\label{app:method-details}

This section specifies the sparse 3D encoder, masking scheme, positional
encoding, and pretraining setup used by our method.

\paragraph{Patch embedding and architecture.}
\label{app:patch-embedding}
The encoder follows the sparse 3D patch-construction and embedding pipeline described in
Section~\ref{sec:method}: point-cloud patches are formed from farthest-point
sampled centers and local nearest-neighbor groups, mapped to continuous patch
embeddings, then processed as sparse 3D tokens by a Transformer encoder with
axis-aware 3D positional encoding. The patch construction follows the
local-patch design of Point-BERT~\cite{yuPointBERTPreTraining3D2022}, but our
patch embedding remains continuous rather than producing discrete codes. We use the same 1024-point
xyz-only observation interface as DP3~\cite{ze3DDiffusionPolicy2024}.
Table~\ref{tab:encoder-config} lists the encoder configuration.

\begin{table}[h]
  \centering
  \small
  \caption{Sparse 3D encoder configuration.}
  \label{tab:encoder-config}
  \begin{tabular}{ll}
    \hline
    Item & Value \\
    \hline
    Input points / channels & 1024 points, xyz only \\
    Patch centers & $M=192$, farthest-point sampling \\
    Local grouping & $k=32$ nearest neighbors per center \\
    Patch coordinates & Relative xyz coordinates \\
    Patch embedding & Two-layer MLP with GELU, max-pooled to one token \\
    Token dimension & $d=384$ \\
    Transformer depth / heads & 6 layers / 8 heads \\
    Transformer MLP ratio & 4 \\
    Output latent dimension & 192 \\
    Encoder parameters & 10.798M \\
    \hline
  \end{tabular}
\end{table}

\paragraph{Masking.}
\label{app:masking}
During pretraining, 70\% of sparse 3D tokens are uniformly sampled and dropped
before the Transformer encoder; the model predicts the alignment action from
the remaining visible tokens. This defines the masked-input condition used in
the pretraining experiments.

\paragraph{3D rotary positional encoding.}
\label{app:rope}
The 3D RoPE module injects patch-center position into attention by splitting
each attention head across the normalized $x$, $y$, and $z$ coordinates. This
keeps positional information tied to workspace coordinates without changing the
sparse tokenization. It extends rotary positional encoding~\cite{suRoFormerEnhancedTransformer2024} to sparse 3D patch centers.
Patch centers are computed after the point-cloud normalization used by the
dataset normalizer. Axial 3D RoPE is applied to queries and keys in every
Transformer layer. With 8 heads and $d=384$, each head has 48 channels; with
RoPE fraction 1.0, the implementation allocates 16 channels to each of the
$x$, $y$, and $z$ axes. The frequency base is 10000.

\paragraph{Pretraining objective and alignment head.}
\label{app:objective-details}
A temporary alignment head maps the encoder latent to the action paired with
each point-cloud observation during pretraining. For LIBERO, the
raw 7D action $(\Delta x,\Delta y,\Delta z,\mathrm{axis\mbox{-}angle},g)$ is
converted to a 10D target
$(\Delta x,\Delta y,\Delta z,\mathrm{rot6d},g)$. For Meta-World, we use the
native 4D action $(\Delta x,\Delta y,\Delta z, g)$. Actions are normalized with the training-set linear
normalizer, including the gripper command. The loss is MSE in normalized action
space. The alignment head is an MLP with hidden widths 256 and 256; downstream
training initializes the encoder and uses its own action decoder.

\paragraph{Pretraining hyperparameters.}
\label{app:pretrain-hparams}
Table~\ref{tab:pretrain-hparams} lists the pretraining hyperparameters. Each
pretraining domain uses one merged zarr dataset, sampled uniformly over
trajectories.

\begin{table}[h]
  \centering
  \small
  \caption{Pretraining hyperparameters.}
  \label{tab:pretrain-hparams}
  \begin{tabular}{ll}
    \hline
    Item & Value \\
    \hline
    Optimizer & AdamW \\
    Learning rate & $2\times10^{-4}$ \\
    Betas / $\epsilon$ & $(0.95,0.999)$ / $10^{-8}$ \\
    Weight decay & $10^{-6}$ \\
    Schedule / warmup & Cosine decay / 500 steps \\
    Batch size & 512 \\
    LIBERO pretraining length & 20{,}000 steps \\
    Meta-World pretraining length & 10{,}000 steps \\
    \hline
  \end{tabular}
\end{table}

\subsection{Simulation Setup}
\label{app:protocol}

This section lists the simulation datasets, preprocessing, downstream training
hyperparameters, and baseline sources.

\paragraph{Simulation benchmarks and data.}
\label{app:sim-data}
We evaluate on LIBERO-10, the 10 long-horizon tasks in the LIBERO suite
\cite{liuLIBEROBenchmarkingKnowledge2023}, and a five-task Meta-World
subset~\cite{yuMetaWorldBenchmarkEvaluation2020}, denoted Meta-World-5, in
simulation. Point-cloud preprocessing follows DP3~\cite{ze3DDiffusionPolicy2024}:
cropped and downsampled to 1024 xyz-only points. Table~\ref{tab:sim-data}
lists the point-cloud preprocessing parameters.

\begin{table}[h]
  \centering
  \small
  \setlength{\tabcolsep}{3pt}
  \caption{Point-cloud preprocessing for simulation benchmarks.}
  \label{tab:sim-data}
  \begin{tabular}{p{0.21\linewidth}p{0.32\linewidth}p{0.32\linewidth}}
    \hline
    Item & LIBERO-10 & Meta-World-5 \\
    \hline
    Cameras &
    agentview, robot0\_eye\_in\_hand &
    corner, gripperPOV \\
    Crop bounds &
    $[-0.5,-1.0,0.2]$ to $[1.0,1.0,1.5]$ &
    $[-0.5,0.4,-0.5]$ to $[0.5,0.95,0.5]$ \\
    Depth range &
    $[0.05,3.0]$ &
    $[0.01,3.0]$ \\
    Voxel size &
    0.008 &
    0.008 \\
    Policy input &
    1024 xyz-only points &
    1024 xyz-only points \\
    \hline
  \end{tabular}
\end{table}

LIBERO-10 uses the released demonstration files; Meta-World-5 uses 50
demonstrations per task generated by the official Meta-World scripted expert
policies.

\begin{figure}[h]
  \centering
  \includegraphics[width=\linewidth]{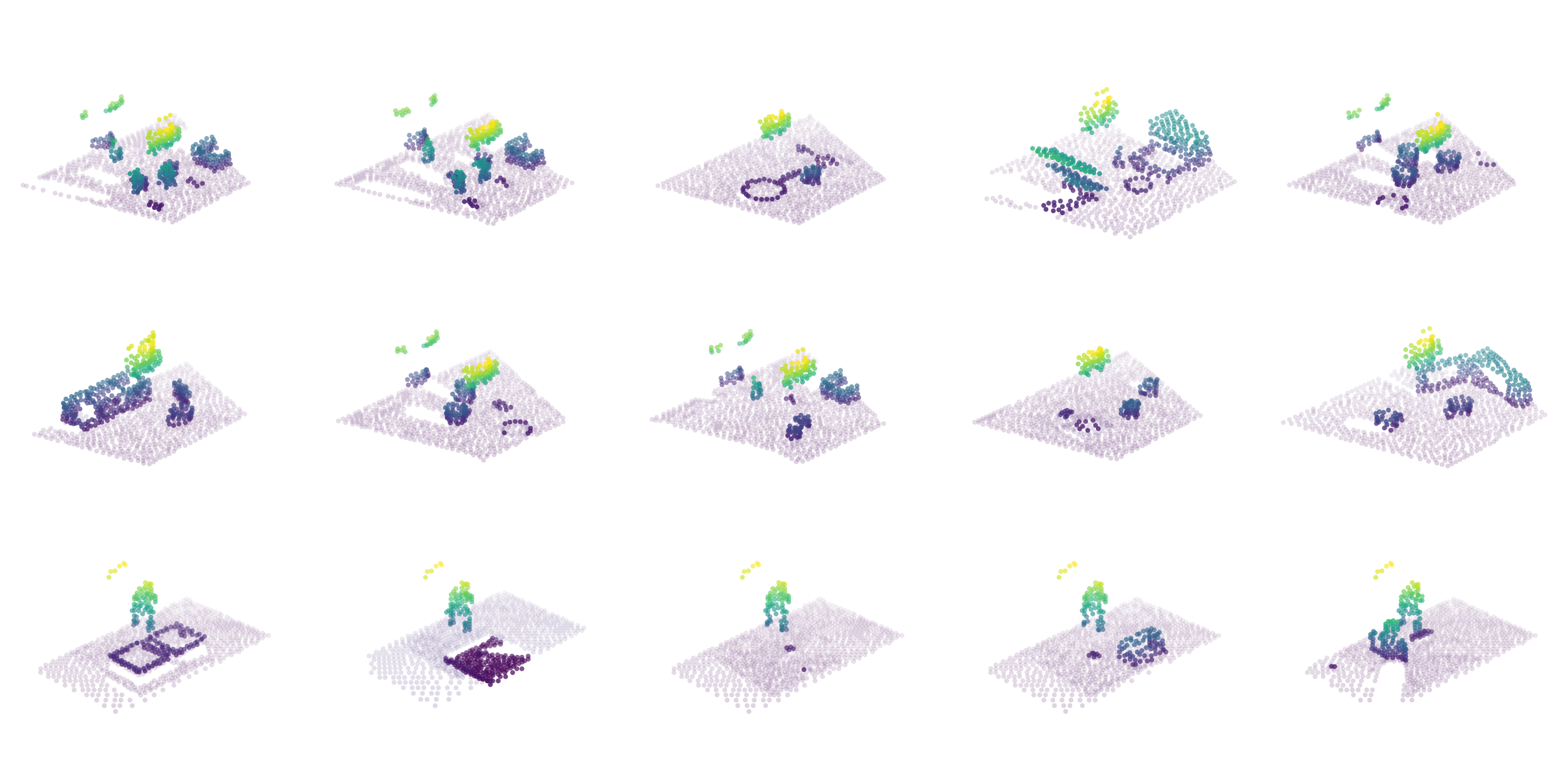}
  \caption{Processed point-cloud observations for all 15 simulation tasks
  (LIBERO-10, rows 1--2; Meta-World-5, row 3), colored by height ($z$-coordinate).
  Each scene is cropped and downsampled to 1024 xyz-only points following the
  preprocessing in Table~\ref{tab:sim-data}. Tabletop points are rendered
  semi-transparent for visualization clarity; no such filtering is applied
  during training.}
  \label{fig:all-tasks-pointcloud}
\end{figure}

\paragraph{Downstream training hyperparameters.}
\label{app:downstream-hparams}
Table~\ref{tab:downstream-hparams} lists the downstream policy hyperparameters.
All downstream policies observe xyz point clouds and robot state.

\begin{table}[h]
  \centering
  \small
  \caption{Downstream fine-tuning hyperparameters.}
  \label{tab:downstream-hparams}
  \begin{tabular}{ll}
    \hline
    Item & Value \\
    \hline
    Optimizer & AdamW \\
    Betas / $\epsilon$ & $(0.95,0.999)$ / $10^{-8}$ \\
    Weight decay & $10^{-6}$ \\
    Learning rate & $5\times10^{-5}$ for the pretrained DP3 fine-tuning preset \\
    Schedule / warmup & Constant / no warmup \\
    Horizon & 16 \\
    Observation steps / action steps & 2 / 8 \\
    Denoising steps & 10 DDIM steps \\
    Batch size & 256 (LIBERO) / 128 (Meta-World) \\
    Train / evaluation seeds & Single training seed; 3 evaluation seeds \\
    \hline
  \end{tabular}
\end{table}

\paragraph{Baseline implementation and provenance.}
\label{app:provenance}
DP3~\cite{ze3DDiffusionPolicy2024} on LIBERO-10 is run by us using the original
implementation adjusted to the LIBERO interface. The reproduced LIBERO-10 DP3
baseline is trained for 20{,}000 downstream steps, evaluated every 2{,}000 steps, and
reported as the average of the top five evaluated checkpoints, following the
original DP3 reporting convention. FVP~\cite{hou4DVisualPretraining}, DP3, and
AFRO~\cite{liangBootstrapDynamicAware3D2025} on Meta-World-5 are literature values taken from the AFRO benchmark
table~\cite{liangBootstrapDynamicAware3D2025}. All columns for our method are
from our runs. SpatialVLA~\cite{quSpatialVLAExploringSpatial2025} and
$\pi_0$~\cite{blackP0VisionLanguageActionFlow2025} are cited from their
original papers and appear only in the main-paper comparison.

\subsection{Additional Simulation Results}
\label{app:per-task}

This section provides full per-task results for the main comparisons and
ablations.

\paragraph{LIBERO-10 in-domain adaptation.}
\label{app:per-task-libero}
Table~\ref{tab:e1-libero10-per-task} reports per-task LIBERO-10 success rates
for our method and DP3 under the in-domain setting.
The rows for our method use the full available LIBERO-10 downstream demonstrations,
LIBERO-10 pretraining, the fixed 500-step checkpoint, and 50 evaluation
episodes per seed across 3 evaluation seeds. DP3 uses the same 1024-point xyz
preprocessing and the same LIBERO task split, but is trained for 20{,}000 downstream
steps and reported as the top-five average over checkpoints evaluated every 2{,}000
steps, matching the original DP3 checkpoint-selection protocol.

\begin{table}[h]
  \centering
  \caption{Per-task LIBERO-10 results. Success rates are reported in
  percentage as mean $\pm$ std.}
  \label{tab:e1-libero10-per-task}
  \begin{tabular}{lcc}
    \hline
    Task & Ours & DP3 \\
    \hline
    \taskname{Soup and Sauce to Basket} & \cellcolor{methodblue}\textbf{83.3 $\pm$ 2.3} & 0.0 $\pm$ 0.0 \\
    \taskname{Cream Cheese and Butter to Basket} & \cellcolor{methodblue}\textbf{88.7 $\pm$ 2.3} & 0.0 $\pm$ 0.0 \\
    \taskname{Turn On Stove and Place Moka Pot} & \cellcolor{methodblue}\textbf{100.0 $\pm$ 0.0} & 100.0 $\pm$ 0.0 \\
    \taskname{Bowl to Bottom Drawer and Close} & \cellcolor{methodblue}\textbf{92.7 $\pm$ 2.3} & 90.0 $\pm$ 0.0 \\
    \taskname{Mugs to Left and Right Plates} & \cellcolor{methodblue}\textbf{84.7 $\pm$ 3.1} & 8.0 $\pm$ 4.0 \\
    \taskname{Book to Back Caddy Compartment} & \cellcolor{methodblue}\textbf{100.0 $\pm$ 0.0} & 0.0 $\pm$ 0.0 \\
    \taskname{Mug to Plate and Pudding Right} & \cellcolor{methodblue}\textbf{75.3 $\pm$ 1.2} & 35.0 $\pm$ 8.4 \\
    \taskname{Soup and Cream Cheese to Basket} & \cellcolor{methodblue}\textbf{76.0 $\pm$ 5.3} & 0.0 $\pm$ 0.0 \\
    \taskname{Both Moka Pots to Stove} & \cellcolor{methodblue}\textbf{78.0 $\pm$ 0.0} & 11.0 $\pm$ 3.7 \\
    \taskname{Mug to Microwave and Close} & \cellcolor{methodblue}\textbf{90.0 $\pm$ 6.0} & 47.0 $\pm$ 7.5 \\
    \hline
    Avg. & \cellcolor{methodblue}\textbf{86.9 $\pm$ 0.7} & 29.1 $\pm$ 36.4 \\
    \hline
  \end{tabular}
\end{table}

\paragraph{Meta-World-5 in-domain and cross-domain adaptation.}
\label{app:per-task-mw}
Table~\ref{tab:metaworld5-full} reports per-task Meta-World-5 success rates
for in-domain Meta-World-5 pretraining and cross-domain LIBERO-10
pretraining, alongside published baseline numbers; see
Appendix~\ref{app:provenance} for column sourcing.
The two columns for our method use identical downstream fine-tuning conditions and
differ only in the pretraining domain. The LIBERO-pretrained encoder is trained
on the merged LIBERO-10 dataset ($10\ \text{tasks} \times 50\ \text{demonstrations}$);
the Meta-World-pretrained encoder is trained on Meta-World-5
($5\ \text{tasks} \times 50\ \text{demonstrations}$). Downstream fine-tuning uses 25
demonstrations per target task. For our Meta-World-5 runs, checkpoints are
evaluated every 400 fine-tuning steps, and each reported value is the mean and
std of the five highest checkpoint success rates within the first 4{,}000 steps for
that task and pretraining source. Each checkpoint uses 20 evaluation episodes
per seed.

\begin{table}[t]
  \centering
  \caption{Per-task Meta-World-5 results under a 25-demonstration budget.}
  \label{tab:metaworld5-full}
  \resizebox{\linewidth}{!}{
  \begin{tabular}{lccccc}
    \hline
    Task & FVP & DP3 & AFRO & Ours (LIBERO pretrain) & Ours (Meta-World pretrain) \\
    \hline
    \taskname{Bin Picking} & 16 & 18 & 20 & \cellcolor{methodblue}63.0 $\pm$ 4.0 & \cellcolor{methodblue}\textbf{90.0 $\pm$ 5.5} \\
    \taskname{Pick Out of Hole} & 26 & 24 & 32 & \cellcolor{methodblue}71.0 $\pm$ 2.0 & \cellcolor{methodblue}\textbf{75.0 $\pm$ 5.5} \\
    \taskname{Push} & 42 & 74 & 78 & \cellcolor{methodblue}91.0 $\pm$ 7.4 & \cellcolor{methodblue}\textbf{95.0 $\pm$ 5.5} \\
    \taskname{Soccer} & 34 & 38 & 36 & \cellcolor{methodblue}54.0 $\pm$ 4.9 & \cellcolor{methodblue}\textbf{72.0 $\pm$ 6.8} \\
    \taskname{Stick Pull} & 24 & 58 & 78 & \cellcolor{methodblue}88.0 $\pm$ 2.5 & \cellcolor{methodblue}\textbf{96.0 $\pm$ 3.7} \\
    \hline
    Avg. & 28.4 & 42.4 & 48.8 & \cellcolor{methodblue}73.4 & \cellcolor{methodblue}\textbf{85.6} \\
    \hline
\end{tabular}}
\end{table}

\paragraph{Pretraining data scaling.}
\label{app:pretrain-data-scaling}
Table~\ref{tab:pretrain-data-scaling-full} gives the per-task values for the
pretraining data-scaling experiment summarized in Figure~\ref{fig:data-efficiency}.
The experiment uses Meta-World-5 with 50 downstream demonstrations per task,
evaluates the fixed 2{,}000-step fine-tuning checkpoint, and uses 20 evaluation
episodes per seed across 3 evaluation seeds. We report two schedules. The
step-scaled schedule uses 1{,}000, 3{,}000, 5{,}000, and 7{,}000 pretraining
steps for 0.1, 0.3, 0.5, and 0.7 of the pretraining data, respectively. The
fixed-step schedule uses 10{,}000 pretraining steps for each subset. The
full-data 10{,}000-step row is the
same masked-input action model used in the objective ablation.

\begin{table}[t]
  \centering
  \small
  \caption{Per-task Meta-World-5 pretraining data scaling. Success rates are
  reported in percentage as mean $\pm$ std at the fixed 2{,}000 fine-tuning
  checkpoint.}
  \label{tab:pretrain-data-scaling-full}
  \resizebox{\linewidth}{!}{
  \begin{tabular}{llcccccc}
    \hline
    Pretraining schedule & Data ratio & \taskname{Bin Picking} & \taskname{Pick Out of Hole} & \taskname{Push} & \taskname{Soccer} & \taskname{Stick Pull} & Avg. \\
    \hline
    Step-scaled & 0.1 & 15.0 $\pm$ 4.1 & 50.0 $\pm$ 7.1 & 95.0 $\pm$ 4.1 & 48.3 $\pm$ 20.5 & 11.7 $\pm$ 8.5 & 44.0 $\pm$ 3.6 \\
    Step-scaled & 0.3 & 76.7 $\pm$ 13.1 & 60.0 $\pm$ 4.1 & 96.7 $\pm$ 4.7 & 33.3 $\pm$ 18.9 & 31.7 $\pm$ 12.5 & 59.7 $\pm$ 8.4 \\
    Step-scaled & 0.5 & 76.7 $\pm$ 4.7 & 61.7 $\pm$ 9.4 & 100.0 $\pm$ 0.0 & 53.3 $\pm$ 20.9 & 33.3 $\pm$ 9.4 & 65.0 $\pm$ 3.6 \\
    Step-scaled & 0.7 & 71.7 $\pm$ 8.5 & 60.0 $\pm$ 7.1 & 100.0 $\pm$ 0.0 & 58.3 $\pm$ 19.3 & 50.0 $\pm$ 10.8 & 68.0 $\pm$ 4.5 \\
    \hline
    Fixed-step & 0.1 & 26.7 $\pm$ 4.7 & 55.0 $\pm$ 7.1 & 100.0 $\pm$ 0.0 & 41.7 $\pm$ 16.5 & 45.0 $\pm$ 10.8 & 53.7 $\pm$ 5.6 \\
    Fixed-step & 0.3 & 75.0 $\pm$ 10.8 & 66.7 $\pm$ 6.2 & 100.0 $\pm$ 0.0 & 55.0 $\pm$ 12.2 & 45.0 $\pm$ 8.2 & 68.3 $\pm$ 6.2 \\
    Fixed-step & 0.5 & 86.7 $\pm$ 8.5 & 51.7 $\pm$ 11.8 & 100.0 $\pm$ 0.0 & 53.3 $\pm$ 17.0 & 58.3 $\pm$ 17.0 & 70.0 $\pm$ 6.2 \\
    Fixed-step & 0.7 & 88.3 $\pm$ 10.3 & 48.3 $\pm$ 10.3 & 100.0 $\pm$ 0.0 & 75.0 $\pm$ 4.1 & 63.3 $\pm$ 13.1 & 75.0 $\pm$ 6.5 \\
    \methodrow
    Fixed-step & 1.0 & 93.3 $\pm$ 7.6 & 58.3 $\pm$ 5.8 & 100.0 $\pm$ 0.0 & 73.3 $\pm$ 20.2 & 85.0 $\pm$ 8.7 & \textbf{82.0 $\pm$ 3.5} \\
    \hline
  \end{tabular}}
\end{table}

Average success improves as more pretraining data is used under both schedules.
At a fixed data ratio, the 10{,}000-step schedule is usually stronger than the
step-scaled schedule, indicating that both data coverage and pretraining
compute affect the downstream representation. The trend is clearest in the
average; individual tasks need not improve monotonically for every subset.

\paragraph{Objective ablation.}
\label{app:per-task-objective}
Table~\ref{tab:e4-objective-ablation-full} gives the per-task breakdown of
the objective ablation summarized in the main paper. The comparison uses
Meta-World-5 with 50 downstream demonstrations per task. All pretrained
variants use 10{,}000 pretraining steps and are evaluated at the fixed 2{,}000
fine-tuning checkpoint with 20 episodes per seed across 3 evaluation seeds.
None trains the downstream policy from scratch. Action only initializes the
encoder from unmasked action alignment. Reconstruction only initializes from
masked sparse 3D token reconstruction: the model predicts the continuous patch
embedding for masked patches and optimizes MSE on masked positions only,
following the masked point modeling family of
objectives~\cite{pangMaskedAutoencodersPoint2022}.
Masked input + action is the main objective, where visible tokens predict the
normalized alignment action.

\begin{table}[t]
  \centering
  \caption{Full per-task objective ablation on Meta-World-5. Success rates
  are reported in percentage as mean $\pm$ std.}
  \label{tab:e4-objective-ablation-full}
  \resizebox{\linewidth}{!}{
  \begin{tabular}{lcccccc}
    \hline
    Pretraining objective & \taskname{Bin Picking} & \taskname{Pick Out of Hole} & \taskname{Push} & \taskname{Soccer} & \taskname{Stick Pull} & Avg. \\
    \hline
    None & 18.3 $\pm$ 7.6 & 53.3 $\pm$ 5.8 & 5.0 $\pm$ 5.0 & 5.0 $\pm$ 0.0 & 13.3 $\pm$ 7.6 & 19.0 $\pm$ 4.6 \\
    None (16{,}000 steps) & 13.3 $\pm$ 5.8 & 50.0 $\pm$ 8.7 & 38.3 $\pm$ 16.1 & 15.0 $\pm$ 8.7 & 25.0 $\pm$ 8.7 & 28.3 $\pm$ 6.7 \\
    Action only & 26.7 $\pm$ 5.8 & 18.3 $\pm$ 5.8 & 98.3 $\pm$ 2.9 & 46.7 $\pm$ 11.5 & 63.3 $\pm$ 14.4 & 50.7 $\pm$ 5.5 \\
    Reconstruction only & 40.0 $\pm$ 8.7 & 46.7 $\pm$ 10.4 & 91.7 $\pm$ 5.8 & 36.7 $\pm$ 28.9 & 61.7 $\pm$ 17.6 & 55.3 $\pm$ 8.0 \\
    \methodrow
    Masked input + action & \textbf{93.3 $\pm$ 7.6} & \textbf{58.3 $\pm$ 5.8} & \textbf{100.0 $\pm$ 0.0} & \textbf{73.3 $\pm$ 20.2} & \textbf{85.0 $\pm$ 8.7} & \textbf{82.0 $\pm$ 3.5} \\
    \hline
  \end{tabular}}
\end{table}

Masked action alignment gives the largest gains on Bin Picking, Soccer,
and Stick Pull. Push reaches high success under the action-only and
reconstruction-only variants.

\paragraph{Action-decoder capacity.}
\label{app:per-task-policy}
Tables~\ref{tab:e6-policy-head-params}
and~\ref{tab:e6-policy-head-simplification-full} report the parameter counts
and per-task results for the action-decoder capacity ablation. The
experiment uses the same fixed 2{,}000-step checkpoint and 20-episode-per-seed setting as
the objective ablation.
The DP3 decoder is the conditional 1D U-Net from DP3~\cite{ze3DDiffusionPolicy2024},
with channel widths 512, 1024, and 2048. The SimpleDP3 decoder uses the same
diffusion interface, condition type, horizon, and 10-step DDIM inference, but
with U-Net widths reduced to 128, 256, and 384. The MLP decoder is a residual MLP
with two 1024-wide hidden layers, GELU activations, LayerNorm, and no dropout.
Parameter counts are computed from instantiated PyTorch models under the
corresponding encoder--decoder configurations.

\begin{table}[t]
  \centering
  \caption{Total parameter count for each encoder--decoder pairing in
  the action-decoder capacity ablation. Parameter counts are reported
  in millions.}
  \label{tab:e6-policy-head-params}
  \begin{tabular}{lccc}
    \hline
    Encoder & DP3 decoder & SimpleDP3 decoder & MLP decoder \\
    \hline
    DP3 encoder & 262.533 & 8.264 & 2.808 \\
    \methodrow
    Pretrained & 273.238 & 18.969 & 13.513 \\
    \hline
  \end{tabular}
\end{table}

\begin{table}[t]
  \centering
  \caption{Full per-task downstream policy-learning results on Meta-World-5.
  Success rates are reported in percentage as mean $\pm$ std.}
  \label{tab:e6-policy-head-simplification-full}
  \resizebox{\linewidth}{!}{
  \begin{tabular}{llcccccc}
    \hline
    Encoder & Action decoder & \taskname{Bin Picking} & \taskname{Pick Out of Hole} & \taskname{Push} & \taskname{Soccer} & \taskname{Stick Pull} & Avg. \\
    \hline
    Scratch (DP3 PointNet) & DP3 decoder & 23.3 $\pm$ 10.4 & 35.0 $\pm$ 10.0 & 8.3 $\pm$ 7.6 & 18.3 $\pm$ 15.3 & 5.0 $\pm$ 5.0 & 18.0 $\pm$ 4.6 \\
    Scratch (DP3 PointNet) & SimpleDP3 decoder & 16.7 $\pm$ 5.8 & 21.7 $\pm$ 25.7 & 8.3 $\pm$ 7.6 & 15.0 $\pm$ 8.7 & 3.3 $\pm$ 2.9 & 13.0 $\pm$ 3.6 \\
    Scratch (DP3 PointNet) & MLP & 15.0 $\pm$ 5.0 & 6.7 $\pm$ 2.9 & 6.7 $\pm$ 7.6 & 11.7 $\pm$ 10.4 & 21.7 $\pm$ 12.6 & 12.3 $\pm$ 5.7 \\
    \methodrow
    Pretrained & DP3 decoder & 93.3 $\pm$ 7.6 & 58.3 $\pm$ 5.8 & \textbf{100.0 $\pm$ 0.0} & \textbf{73.3 $\pm$ 20.2} & \textbf{85.0 $\pm$ 8.7} & \textbf{82.0 $\pm$ 3.5} \\
    \methodrow
    Pretrained & SimpleDP3 decoder & \textbf{98.3 $\pm$ 2.9} & 56.7 $\pm$ 7.6 & 98.3 $\pm$ 2.9 & 68.3 $\pm$ 15.3 & 53.3 $\pm$ 18.9 & 75.0 $\pm$ 8.9 \\
    \methodrow
    Pretrained & MLP & 88.3 $\pm$ 2.9 & \textbf{70.0 $\pm$ 13.2} & \textbf{100.0 $\pm$ 0.0} & 70.0 $\pm$ 8.7 & 45.0 $\pm$ 0.0 & 74.7 $\pm$ 3.5 \\
    \hline
  \end{tabular}}
\end{table}

The pretrained encoder improves all three action decoders over their scratch
counterparts. The best decoder varies by task: SimpleDP3 is strongest on Bin
Picking, the MLP is strongest on Pick Out of Hole, and the full DP3 decoder is
strongest on Stick Pull.

\paragraph{Model scale.}
\label{app:scale}
The DP3 PointNet encoder has 0.092M parameters, while our sparse patch
encoder has 10.798M parameters. The scratch rows in
Table~\ref{tab:e6-policy-head-simplification-full} use the same 2{,}000-step
training schedule, single training seed, and 3 evaluation seeds as the pretrained rows.

\subsection{Real-Robot Setup and Evaluation}
\label{app:real-robot}

\paragraph{Platform and tasks.}
We use a single-arm AgileX PiPER 6-DoF robotic arm with a fixed Intel
RealSense D415 depth camera.
Four tasks are evaluated: \taskname{Bin Picking}, \taskname{Cube Stacking},
\taskname{Mug \& Plate}, and \taskname{Mug \& Toy}, with 50 real
demonstrations per task. We additionally
build calibrated, task-matched simulated scenes by replicating the robot model,
workspace geometry, objects, and task layouts, and collect 100 simulated
demonstrations per task.

\begin{figure}[H]
  \centering
  \includegraphics[width=\linewidth]{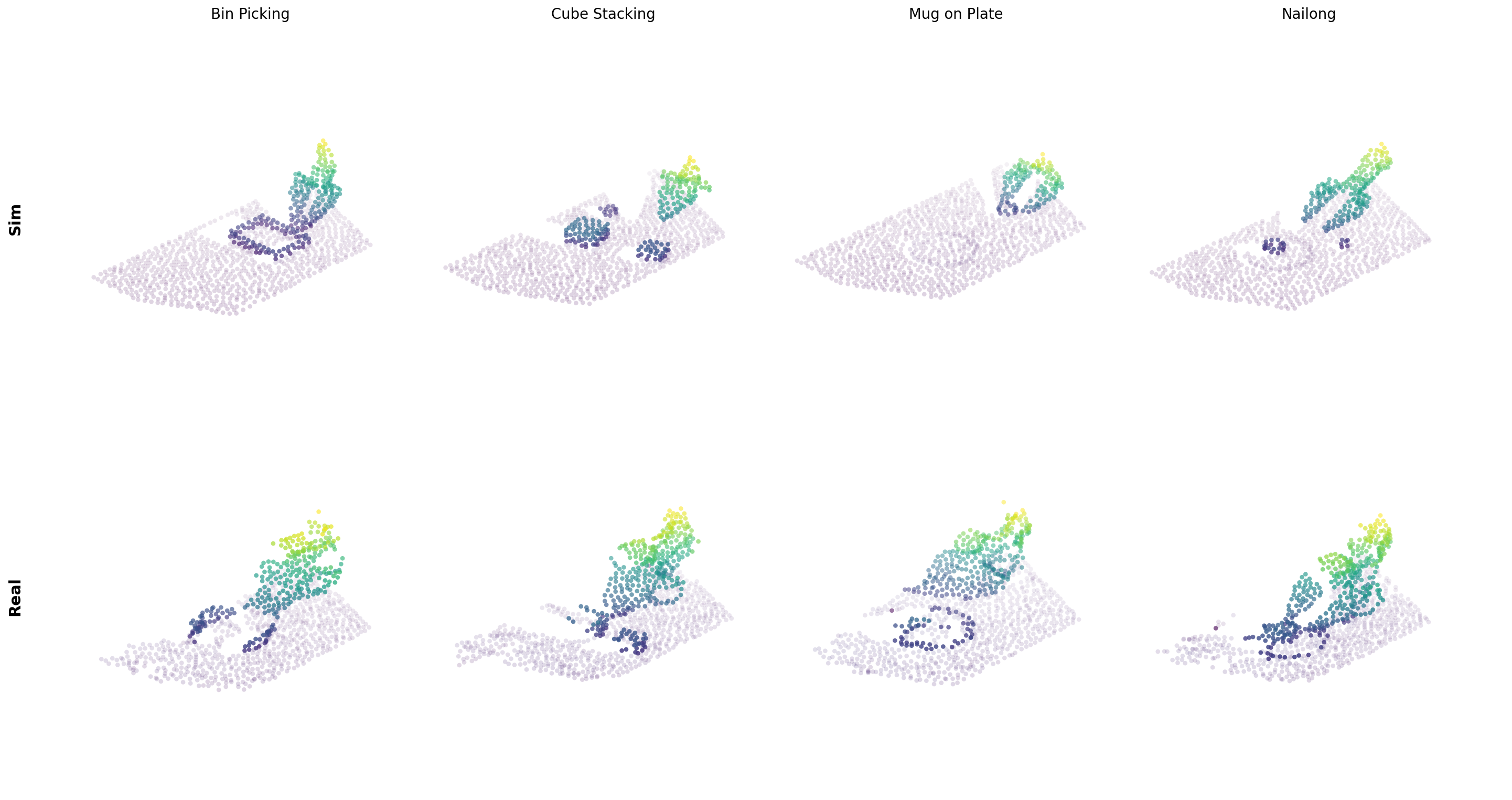}
  \caption{Processed sim and real point-cloud observations for the four
  PiPER evaluation tasks. Each column shows one task, and the two rows show the
  corresponding task-matched simulation and real-robot observations after the
  shared point-cloud preprocessing pipeline. Points are colored by height for
  visualization.}
  \label{fig:real-sim-pointclouds}
\end{figure}

\paragraph{Simulated demonstration collection.}
To improve position generalization, simulated demonstrations randomize object
and distractor positions within a workspace that is slightly larger than the
real-robot task region, while preserving the task semantics and calibrated
robot setup used for real-robot fine-tuning.

\begin{figure}[H]
  \centering
  \includegraphics[width=\linewidth]{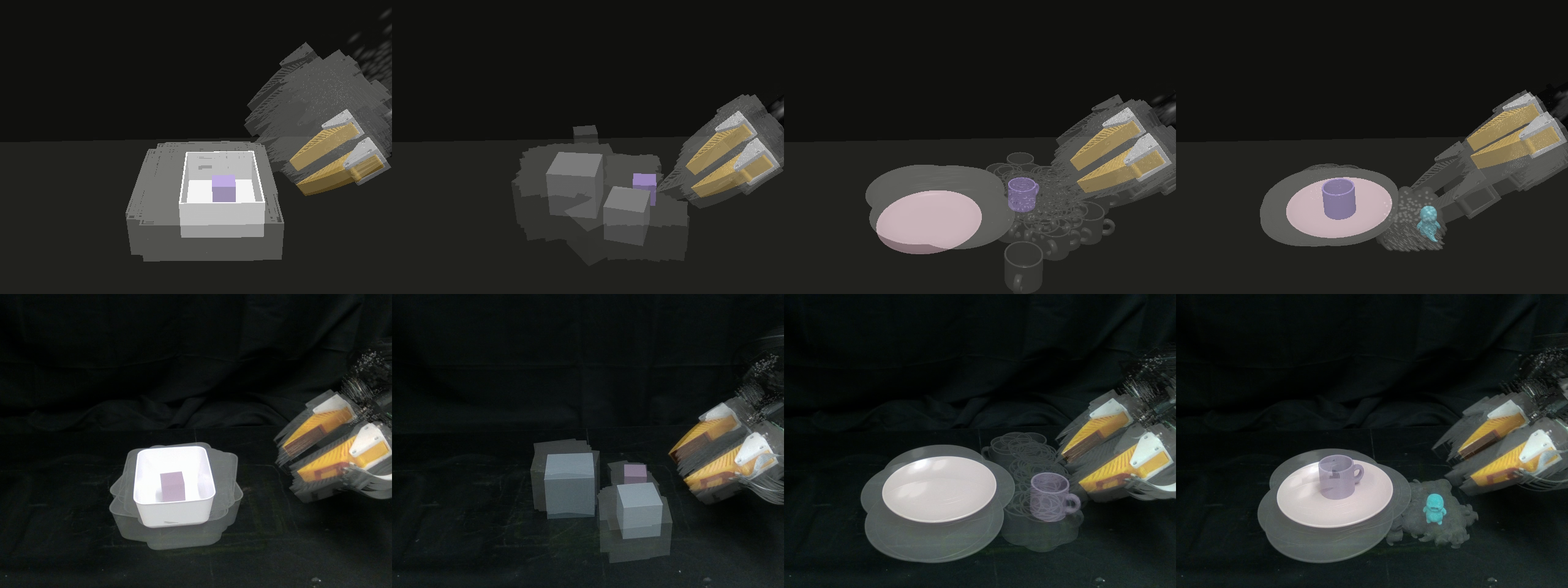}
  \caption{Examples of the position-randomized simulated demonstrations used
  for real-robot pretraining. The simulated collection varies object and
  distractor placement across the four target tasks in a slightly larger
  workspace than the real-robot task region, while keeping the same robot setup
  and task structure.}
  \label{fig:real-position-randomization}
\end{figure}

\paragraph{Real-robot training protocols.}
The real-robot comparison uses three training conditions. The direct
mixed-data baselines train a target-task policy from scratch on the matched
100 simulated and 50 real demonstrations for that task, using either the DP3
policy or our downstream policy architecture. The pretrained variant first
pretrains the encoder on the 100 simulated demonstrations for each target task
(10{,}000 pretraining steps), then fine-tunes on the same 50 real demonstrations.
The pretrained variant uses the downstream fine-tuning setting in
Table~\ref{tab:downstream-hparams}, with batch size 128, and reports the fixed
4{,}000-step fine-tuning checkpoint. The direct DP3 baseline is evaluated at
5{,}000, 10{,}000, 15{,}000, and 20{,}000 training steps, and we report its
best checkpoint.

\begin{table}[H]
  \centering
  \small
  \caption{Training protocols and data sources for the real-robot deployment
  results in Table~\ref{tab:real-robot}.}
  \label{tab:real-robot-protocols}
  \resizebox{\linewidth}{!}{
  \begin{tabular}{lllll}
    \hline
    Method & Pretrain source / steps & Fine-tune source & Fine-tune budget & Reported checkpoint \\
    \hline
    DP3 & -- & 100 sim + 50 real/task & 20{,}000 steps & Best of 5{,}000, 10{,}000, 15{,}000, 20{,}000 \\
    Ours from scratch & -- & 100 sim + 50 real/task & 4{,}000 steps, batch 128 & 4{,}000 fixed checkpoint \\
    \methodrow
    Ours & 100 sim/task, 10{,}000 steps & 50 real/task & 4{,}000 steps, batch 128 & 4{,}000 fixed checkpoint \\
    \hline
  \end{tabular}}
\end{table}

\paragraph{Observation and action interface.}
Point clouds are cropped to 1024 xyz-only points in the world frame centered
at the table surface. Pretraining uses end-effector delta pose with absolute
gripper width, matching the simulation interface. Fine-tuning uses delta joint
positions ($\Delta q_1,\ldots,\Delta q_6$) with absolute gripper commands for
the real-robot controller, while keeping the same encoder initialization.

\paragraph{Success criteria.}
We count a rollout as successful only when all required task stages are
completed in order. For \taskname{Bin Picking}, the robot must approach the
target object, grasp it, lift it, and place it at the target location. For
\taskname{Mug \& Plate}, the robot must approach the mug, grasp it, lift it,
and place it inside the plate. For \taskname{Cube Stacking}, the gripper must
close, push the medium cube next to the large cube, approach the small cube,
grasp the small cube, lift it, and place it on top of the large cube. For
\taskname{Mug \& Toy}, the robot must approach and grasp the mug, place the
mug aside, approach the toy, grasp the toy, and place the toy inside the plate.

\paragraph{Qualitative failure modes.}
The main real-robot failures occur during precise approach, lifting, placement,
or multi-stage recovery. In \taskname{Bin Picking}, the gripper sometimes
approaches the bin at insufficient height or lifts the grasped object without
enough clearance. In \taskname{Mug \& Plate}, the mug is sometimes not lifted
high enough before moving over the plate during placement. In
\taskname{Cube Stacking}, failures include out-of-distribution states during
the pushing stage and insufficient lifting height before the stacking
placement. In \taskname{Mug \& Toy}, failures include empty grasps on the toy
and out-of-distribution states during the multi-stage sequence. These cases
motivate future extensions with temporal context and recovery-oriented data.

\begin{figure}[H]
  \centering
\includegraphics[width=\linewidth]{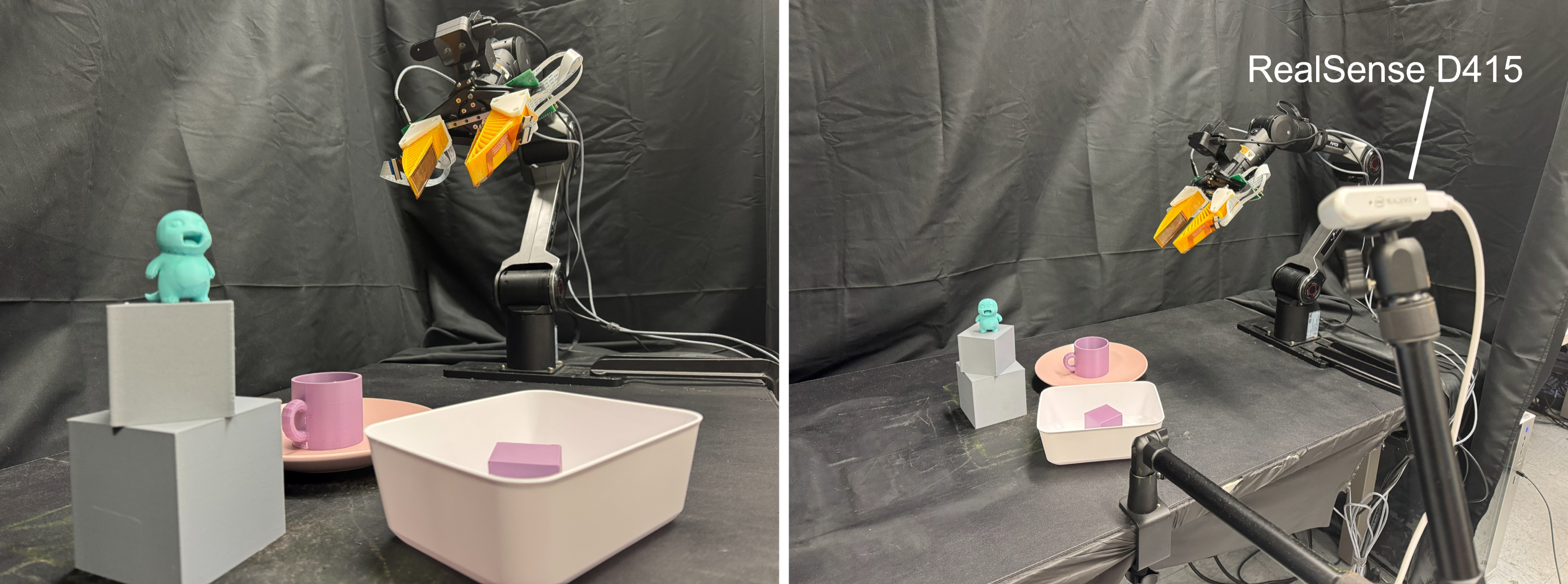}
  \caption{Real-robot experimental setup. Left: the PiPER arm with the object
  set used across all four evaluation tasks. Right: the mounted Intel RealSense D415
  depth camera used to capture point-cloud observations.}
  \label{fig:real-robot-setup}
\end{figure}

\begin{figure}[H]
  \centering
  \includegraphics[width=\linewidth]{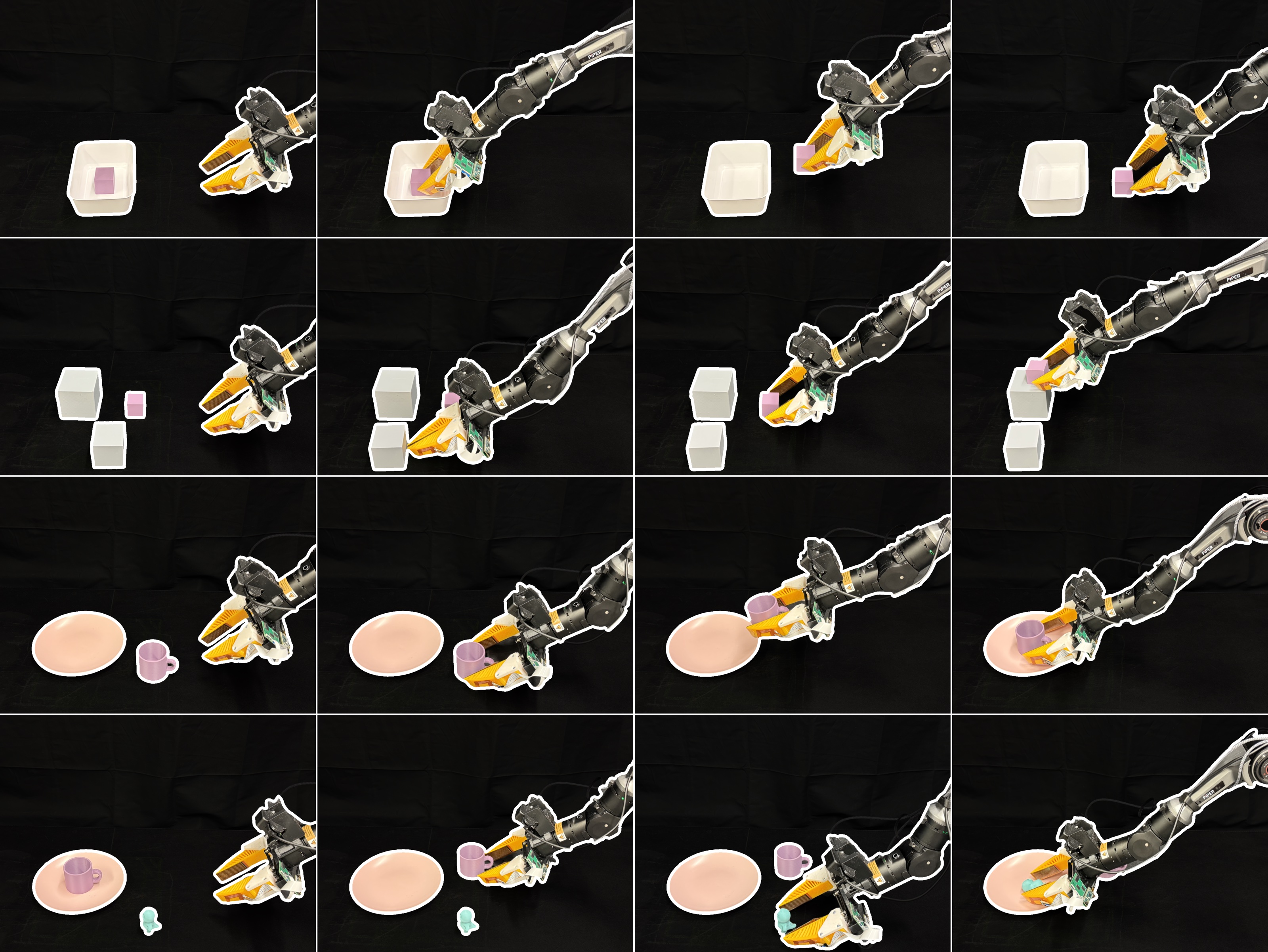}
  \caption{Representative real-robot task completion sequences for the four
  PiPER evaluation tasks. Each row shows a rollout progressing from the initial
  scene to successful task completion.}
  \label{fig:real-robot-completion-sequences}
\end{figure}

\end{document}